\documentclass[lettersize,journal]{IEEEtran}
\usepackage{amsmath,amsfonts}
\usepackage{algorithmic}
\usepackage{algorithm}
\usepackage{array}
\usepackage[caption=false,font=normalsize,labelfont=sf,textfont=sf]{subfig}
\usepackage{textcomp}
\usepackage{stfloats}
\usepackage{url}
\usepackage{verbatim}
\usepackage{graphicx}
\usepackage{cite}

\usepackage{dsfont}
\usepackage{multirow}
\usepackage{hyperref}
\usepackage{tabularx}
\usepackage{amssymb}

\begin{document}

\title{Low-Contrast-Enhanced Contrastive Learning for Semi-Supervised Endoscopic Image Segmentation}

\author{Lingcong Cai, Yun Li, Xiaomao Fan, Kaixuan Song, Ruxin Wang, and Wenbin Lei 
\thanks{This work is partially supported by the National Natural Science Foundation of China (62473267), the Basic and Applied Basic Research Project of Guangdong Province (2022B1515130009), the Special subject on Agriculture and Social Development, Key Research and Development Plan in Guangzhou (2023B03J0172), and the Natural Science Foundation of Top Talent of SZTU (GDRC202318).}
\thanks{Lingcong Cai, Xiaomao Fan, and Kaixuan Song are with the College of Big Data and Internet, Shenzhen Technology University, Shenzhen 518118, China (e-mail: cailingcong@gmail.com; astrofan2008@gmail.com; lki9988@163.com).  }
\thanks{Yun Li and Wenbin Lei are with The First Affiliated Hospital, Sun Yat-Sen University, Guangzhou, China (e-mail: liyun76@mail.sysu.edu.cn; leiwb@mail.sysu.edu.cn).}
\thanks{Ruxin Wang is with the Shenzhen Institute of Advanced Technology, Chinese Academy of Sciences, China (e-mail: rx.wang@siat.ac.cn).}
\thanks{Lingcong Cai and Yun Li are co-first authors.}
\thanks{Xiaomao Fan and Wenbin Lei are corresponding authors.}}

\markboth{Journal of \LaTeX\ Class Files,~Vol.~14, No.~8, August~2021}%
{Shell \MakeLowercase{\textit{et al.}}: A Sample Article Using IEEEtran.cls for IEEE Journals}


\maketitle

\begin{abstract}
The segmentation of endoscopic images plays a vital role in computer-aided diagnosis and treatment. The advancements in deep learning have led to the employment of numerous models for endoscopic tumor segmentation, achieving promising segmentation performance. Despite recent advancements, precise segmentation remains challenging due to limited annotations and the issue of low contrast. To address these issues, we propose a novel semi-supervised segmentation framework termed LoCo via low-contrast-enhanced contrastive learning (LCC). This innovative approach effectively harnesses the vast amounts of unlabeled data available for endoscopic image segmentation, improving both accuracy and robustness in the segmentation process. Specifically, LCC incorporates two advanced strategies to enhance the distinctiveness of low-contrast pixels: inter-class contrast enhancement (ICE) and boundary contrast enhancement (BCE), enabling models to segment low-contrast pixels among malignant tumors, benign tumors, and normal tissues. Additionally, a confidence-based dynamic filter (CDF) is designed for pseudo-label selection, enhancing the utilization of generated pseudo-labels for unlabeled data with a specific focus on minority classes. Extensive experiments conducted on two public datasets, as well as a large proprietary dataset collected over three years, demonstrate that LoCo achieves state-of-the-art results, significantly outperforming previous methods. The source code of LoCo is available at the URL of \href{https://github.com/AnoK3111/LoCo}{https://github.com/AnoK3111/LoCo}.
\end{abstract}


\section{Introduction}
\label{sec:introduction}
\IEEEPARstart{E}{lectronic} endoscope is indispensable for diagnosing tumors in the throat and digestive tract, offering exceptionally high clarity, resolution, and vivid color imaging to detect early disease signs—especially the small, concealed lesions often associated with malignant tumors \cite{cadamuro2023advanced}. However, current tumor detection and diagnostic accuracy are constrained by the physician’s expertise and experience level. Precisely segmenting tumor lesions can yield valuable annotated information on affected areas, as well as refined analyses of lesion size, shape, and color \cite{jiang2023review}. This capability is essential for accurately identifying and diagnosing tumor abnormalities, enabling more effective tumor screening, personalized treatment planning, and ultimately enhancing patient outcomes.

\begin{figure}[tb]
\centerline{\includegraphics[width=\columnwidth]{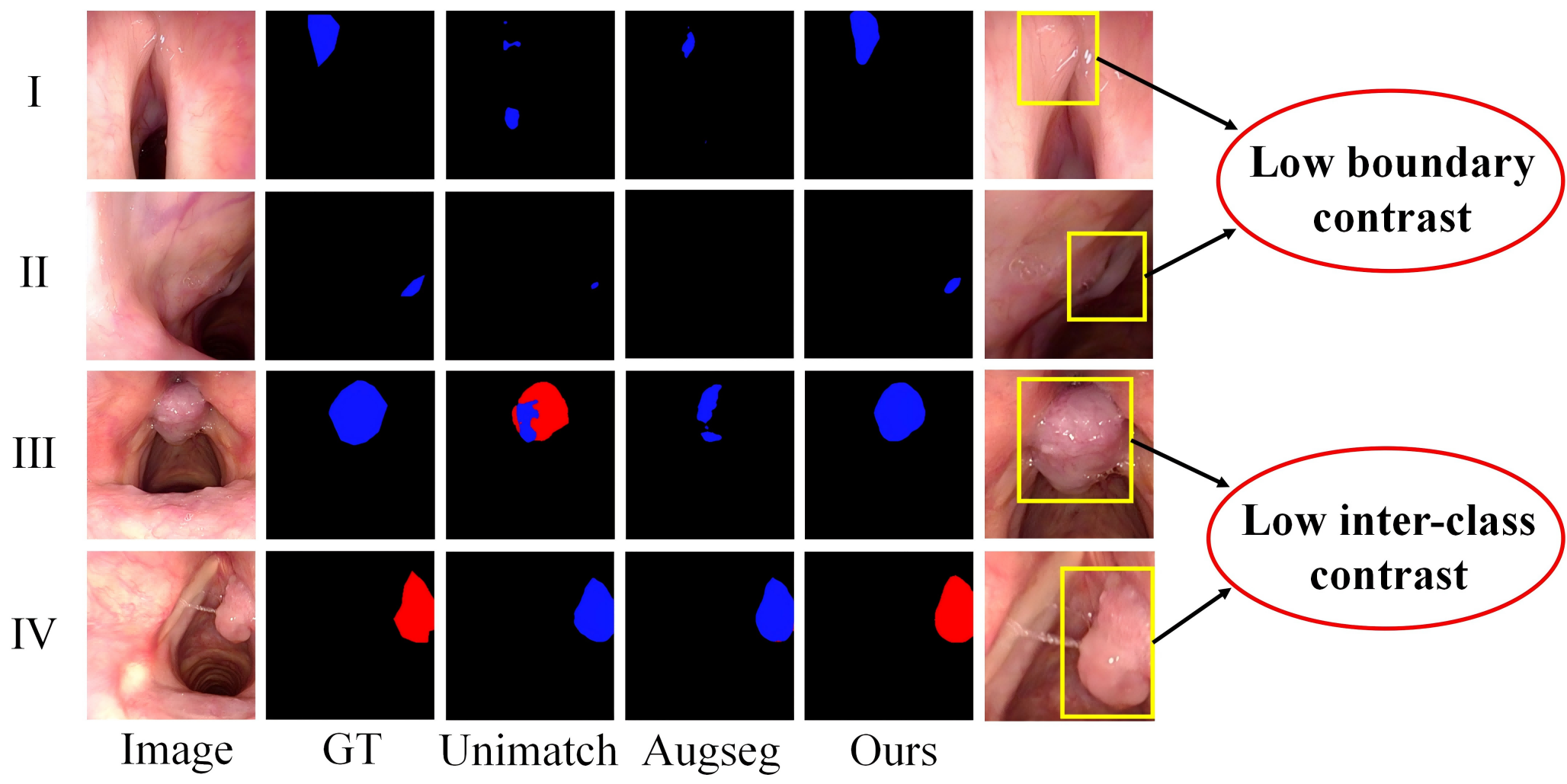}}
\caption{Illustration of low-contrast pixels in endoscopic images. Rows \(\mathrm{I}\) and \(\mathrm{II}\) illustrate the low boundary contrast, which refers to the low contrast between tumor and normal tissues. Rows \(\mathrm{III}\) and \(\mathrm{IV}\) illustrate the low inter-class contrast, which refers to the low contrast between benign and malignant tumors.}
\label{Figure:introduction}
\end{figure}

Recently, remarkable advancements in deep learning models have led to notable improvements in semantic segmentation across diverse domains, yielding promising outcomes \cite{ chen2018encoder, unet++, lvit, h2former,zhou2024uncertainty,li2023erdunet}. However, these methods generally rely on extensive manual annotations to attain competitive performance. In endoscopic applications, the limitation of the annotations constrains model performance on endoscopic image datasets. Semi-supervised learning offers an efficient solution by leveraging limited labeled data with abundant unlabeled data for training. Current segmentation methods \cite{wang2023hunting, zhao2023augmentation, yang2023revisiting, sun2024corrmatch} primarily integrate consistency learning \cite{sohn2020fixmatch} with pseudo-labeling \cite{lee2013pseudo}. These methods utilize model predictions on weakly perturbed unlabeled data as pseudo-labels to supervise predictions on strongly perturbed data, while applying a confidence threshold to filter out low-confidence pseudo-labels, thus reducing confirmation bias \cite{arazo2020pseudo}. Additionally, segmentation foundation models, such as MedSAM \cite{MedSAM}, offer an effective approach to mitigating the challenge of limited annotations. This is due to their strong generalization capabilities, which allow them to perform well on downstream tasks with only a small amount of data for fine-tuning. Despite these advances, the issue of low contrast in endoscopic images still poses a significant challenge for the model to accurately segment regions corresponding to normal, benign, and malignant tissues.

The issue of low contrast can be classified into two categories: low boundary contrast and low inter-class contrast (see Fig. \ref{Figure:introduction}). Low boundary contrast arises from the high similarity between normal and tumor tissues, compounded by the blurred boundaries of tumors. In contrast, low inter-class contrast is characterized by the high similarity among different tumor types. To address low boundary contrast, several studies have incorporated boundary information to enhance the model's capacity to differentiate between normal and tumor regions. Notable approaches include PraNet \cite{fan2020pranet}, BCNet \cite{yue2022boundary}, and ACL-Net \cite{wu2023acl}. For example, Polyper \cite{shao2024polyper} introduced boundary-sensitive attention mechanisms, improving segmentation accuracy by refining features in proximity to boundary regions. Conversely, there is limited research focused on low inter-class contrast. This scarcity is largely due to the fact that most existing endoscopic datasets feature only two categories: background and tumor, lacking detailed annotations for various tumor types. Consequently, related studies face challenges in effectively addressing the issue of low inter-class contrast. To bridge the gap, this work utilizes a large proprietary dataset collected from the First Affiliated Hospital, Sun Yat-sen University. This dataset includes pixel-level annotations for three categories: normal tissue, benign tumors, and malignant tumors, thereby facilitating research aimed at tackling the challenge of low inter-class contrast.

In this study, we propose a novel semi-supervised segmentation framework, termed LoCo, which employs low-contrast-enhanced contrastive learning (LCC) for endoscopic image segmentation. Specifically, to facilitate the model's learning of discriminative features, we propose two strategies: inter-class contrast enhancement (ICE) and boundary contrast enhancement (BCE) for selecting low-contrast pixels to construct positive and negative pairs for contrastive learning. In the ICE, we identify low-contrast pixels that exhibit the lowest similarity to class embeddings within each category for contrastive learning, enabling the model to establish a more robust decision boundary for effectively differentiating between various tumor types. In the BCE, we identify low-contrast boundary pixels with the highest boundary feature similarity for contrastive learning, which enhances the distinct separation of tumor boundaries from normal tissue. Furthermore, to tackle the challenges of low utilization of pseudo-labels, we design a confidence-based dynamic filter (CDF). This filter dynamically adjusts thresholds for different categories based on the class-wise confidence of pseudo-labels, thereby enhancing the utilization of pseudo-labels for minority classes. Experimental results from a proprietary laryngeal cancer dataset and two polyp segmentation datasets indicate that our proposed LoCo framework achieves state-of-the-art performance, exceeding existing methods by a significant margin. The primary contributions of our work can be summarized as follows:

\begin{itemize}
    \item We propose a novel semi-supervised framework for endoscopic image segmentation, applicable to a range of segmentation tasks in endoscopy, including laryngeal cancer and polyp segmentation. 
    \item We develop two advanced strategies to enhance the distinctiveness of low-contrast pixels: inter-class contrast enhancement and boundary contrast enhancement. These approaches enable models to effectively capture the discriminative features that differentiate malignant tumors, benign tumors, and normal tissues.
    \item We design a confidence-based dynamic filter to improve the pseudo-label utilization of minority classes.
    \item Experiments conducted on a proprietary laryngeal cancer dataset and two polyp segmentation datasets show that our proposed framework can achieve a new state-of-the-art result in semi-supervised endoscopic image segmentation.
\end{itemize}

The remainder of the paper is organized as follows. Section \ref{section:related work} provides a brief review of related work pertinent to this article. Section \ref{section:method} details the proposed LoCo method. Section \ref{section:experiments} presents and analyzes the relevant experimental results. Finally, Section \ref{section:conclusion} reviews and summarizes the article.

\section{Related Works}
\label{section:related work}

\subsection{Semi-supervised semantic segmentation}
Semi-supervised semantic segmentation seeks to leverage the abundant visual information found in unlabeled images to enhance segmentation accuracy \cite{ran2024pseudo}. Recently, these approaches have made significant strides in fields such as medical image analysis \cite{lei2022semi, shen2023co, PLGCL, miao2023caussl, chi2024adaptive} and remote sensing interpretation \cite{bandara2022revisiting, zhang2023text2seg, yuan2024dynamically}. Most studies draw inspiration from advancements in semi-supervised learning, proposing straightforward yet effective designs centered around entropy minimization \cite{xie2020self, zoph2020rethinking, pham2021meta} and consistency regularization \cite{sohn2020fixmatch, zhao2022lassl}. Entropy minimization involves assigning pseudo-labels to unlabeled data by utilizing the knowledge derived from labeled samples, followed by retraining the model on the combined dataset. This approach is exemplified by methods such as CPS \cite{chen2021semi}, CCVC \cite{wang2023conflict}, and St++ \cite{yang2022st++}. Despite its effectiveness, the offline self-training pipeline is rarely implemented due to the complexities associated with its three-stage training process. As a result, many studies prefer end-to-end pseudo-labeling frameworks driven by consistency regularization, which focus on generating perturbations in input data and require the model to produce consistent predictions for the same data across various augmentations. Notable examples include AugSeg \cite{zhao2023augmentation}, UniMatch \cite{yang2023revisiting}, and DDFP \cite{wang2024towards}. However, these methods often overlook the issue of low contrast, which can lead to suboptimal segmentation performance, particularly in endoscopic images.

\subsection{Contrastive learning}
Contrastive learning is a powerful approach for representation learning, designed to enhance the model's ability to generate robust representations by minimizing the distance between positive pairs while maximizing the distance between negative pairs. This technique has shown remarkable success in both self-supervised \cite{moco,simclr,byol,SwAV,Barlow_twins} and supervised \cite{supcon} learning contexts. Recently, contrastive learning has been introduced to segmentation tasks, with methods such as P2Seg \cite{P2Seg}, C3-SemiSeg \cite{C3-semiseg}, U$^{2}$PL \cite{wang2022semi}, and PLGCL \cite{PLGCL} utilizing this framework to extract more effective supervisory signals, thereby enabling the model to learn more discriminative features. While these advancements have led to improved segmentation performance, some studies \cite{wang2023hunting,HPC} indicate that the careful selection of samples—particularly those at the edges of the distribution—for constructing positive and negative pairs is critical for learning clearer decision boundaries. In our proposed low-contrast-enhanced contrastive learning approach, we specifically select low-contrast pixels from both categorical and boundary perspectives to form positive and negative pairs. We then employ a novel low-contrast-enhanced contrastive loss to optimize the distances of these pairs, facilitating the model's ability to learn more pronounced contrastive features in endoscopic images.

\begin{figure*}[!t]
\centerline{\includegraphics[width=2.0\columnwidth]{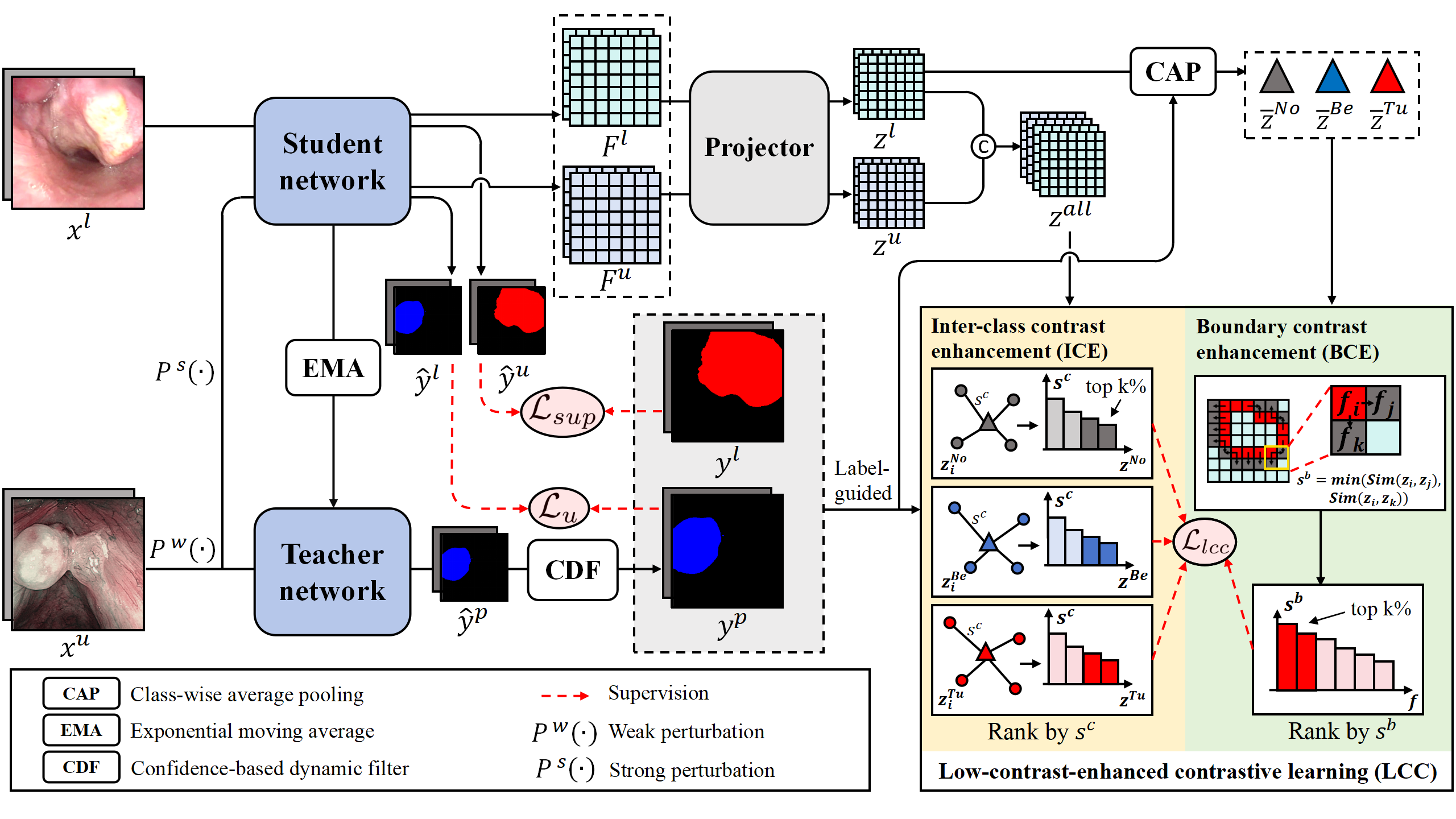}}
\caption{The overall architecture of LoCo. It is designed based on the mean-teacher framework, featuring two branches: a student network and a teacher network. The teacher network generates reliable pseudo-labels using a confidence-based dynamic filter (CDF). These pseudo-labels, along with labeled images, are combined to supervise the learning of the student network. Additionally, low-contrast-enhanced contrastive learning (LCC) is employed to improve segmentation performance in low-contrast images through inter-class contrast enhancement (ICE) and boundary contrast enhancement (BCE).}
\label{framework}
\end{figure*}

\section{Methods}
\label{section:method}
\subsection{Overall LoCo}
Fig.\ref{framework} illustrates the overall architecture of LoCo, which is based on the mean-teacher framework. The LoCo comprises three primary components: a student network, a teacher network, and a Low-contrast-enhanced contrastive learning module (\emph{i.e.}, LCC, see Section \ref{sec LCC}). For the teacher network, it utilizes the same architecture as the student network, updating its parameters via an Exponential Moving Average (EMA) scheme. The EMA is defined as follows:
\begin{equation}
    \theta_{t} \leftarrow \alpha \theta_{t} + (1 - \alpha) \theta_{s},
\end{equation}
\noindent
where $\theta_{s}$ and $\theta_{t}$ represent the learned parameters of the student network and the teacher network, respectively. $\alpha \in [0, 1]$ is a momentum hyperparameter, and it is typically set to 0.999 by default \cite{tarvainen2017mean, zhao2023augmentation}. Regarding the student network, its parameters $\theta_{s}$ are updated with the optimization of Stochastic Gradient Descent (SGD). In the medical domain, the availability of labeled medical images is often limited, while the volume of unlabeled images can be substantial. This scarcity of labeled data hampers the model's ability to achieve satisfactory segmentation results. Consequently, effectively utilizing the abundant unlabeled data is crucial for training a robust model. In this study, we utilize the teacher network to generate reliable pseudo-labels through a Confidence-based Dynamic Filter (\emph{i.e.}, CDF, see Section \ref{sec CDF}) for unlabeled data. These pseudo-labels, along with labeled images, are combined to supervise the learning of the student network. Additionally, LCC is employed to improve segmentation performance in low-contrast images through inter-class contrast enhancement (\emph{i.e.}, ICE, see Section \ref{sec ICE}) and boundary contrast enhancement (\emph{i.e.}, BCE, see Section \ref{sec BCE}).

Specifically, consider a batch of labeled set $B^{l}=\{x_{i}^{l},y_{i}^{l}\}_{i=1}^{N^l}$ and a batch of unlabeled set $B^{u}=\{x_{i}^{u}\}_{i=1}^{N^u}$. Here, $N^l$ and $N^u$ indicate the number of labeled data and unlabeled data, respectively. $y_i^l$ represents the label for the corresponding labeled data $x_i^l$. Specifically, $y_{i,j}^l\in\{1, 2,\cdots,K\}$ is the label of the $i$-th image at the $j$-th position, where $K$ denotes the number of classes. The objective of LoCo is to maximize the supervision derived from the limited labeled data while effectively utilizing the unlabeled data to enhance model training. 

For a labeled data $x_{i}^{l} \in \mathbb{R}^{H \times W \times D}$, we employ a standard pixel-wise cross-entropy loss $\mathcal{L}_{ce}$ to supervise pixel classification, where $H \times W$ represents the image resolution and $D$ stands for the number of channels. The supervision loss for labeled data, represented as $\mathcal{L}_{sup}$, can be defined as:
\begin{equation}
\mathcal{L}_{sup}=\frac{1}{N^l}\sum_{i=1}^{N^{l}}\frac{1}{H \times W}\sum_{j=1}^{H \times W}\mathcal{L}_{ce}(\hat{y}_{i,j}^{l}, y_{i,j}^{l}),
\end{equation}
\noindent
where $\hat{y}_{i}^{l}=f_{s}(x_{i}^{l};\theta_{s}) \in \mathbb{R}^{H \times W \times K}$ denotes the predicted probability of the student network on the $i$-th labeled data and $f_{s}(\cdot)$ represents the student network.

For an unlabeled data $x_{i}^{u} \in \mathbb{R}^{H \times W \times D}$, we adopt consistency regularization to mine effective supervision. Following most existing work \cite{french2020semi,zhao2023augmentation,yang2023revisiting,sun2024corrmatch}, weak perturbation $\mathcal{P}^{w}(\cdot)$ and strong perturbation $\mathcal{P}^{s}(\cdot)$ are first applied to generate the prediction disagreement. Here, weak perturbation is a series of geometrical augmentations, including random scale, random scale, and random flip. Strong perturbation consists of some intensity-based augmentation, such as Gaussian blur and color jitter, and some layout changes like CutMix \cite{french2020semi}. For the student network, an unlabeled image undergoes a weak perturbation, followed by a strong perturbation, producing a new augmented image. This new image is then fed into the student network to generate the prediction $\hat{y}_{i}^{u}$, which is defined as:
\begin{equation}
    \hat{y}_{i}^{u}=f_{s}(\mathcal{P}^{s}(\mathcal{P}^{w}(x_{i}^{u}));\theta_{s}).
\end{equation}
For the teacher network, after processing by a weak perturbation, an unlabeled image is fed into the teacher network to generate the prediction $\hat{y}_{i}^{p}$, which is defined as:
\begin{equation}
    \hat{y}_{i}^{p}=f_{t}(\mathcal{P}^{w}(x_{i}^{u});\theta_{t}),
\end{equation}
\noindent where $f_{t}(\cdot)$ denotes the teacher network. Then, to prevent confirmation bias and improve the utilization of pseudo-labels, we introduce a CDF module to screen out those unreliable predictions to generate reliable pixel-level pseudo-labels $y_{i}^{p}$. Specifically, for each class $c$, a dynamic threshold $\mathcal{T}(c)$ is employed to filter out those low-confidence pseudo-labels belonging to this class. Then, an $\arg\max$ operation is applied to obtain the hard pseudo-labels $y^p$, formulated as:
\begin{equation}
y_{i,j}^{p}=\left\{
	\begin{aligned}
	  & p_{i,j}, &\ \max(\hat{y}_{i,j}^{p})>=\mathcal{T}(p_{i,j}),\\
	& \varnothing, &\ otherwise,\\
	\end{aligned}
        \right
        .
\end{equation}

\noindent where $p_{i,j}=\arg\max(\hat{y}_{i,j}^{p})$ and $\varnothing$ represents the filtered pseudo-label.

Finally, the pixel-level cross-entropy $\mathcal{L}_{ce}$ is utilized to enforce the consistency between the teacher and student network. The unsupervised loss $\mathcal{L}_{u}$ for unlabeled data with pseudo-labels can be calculated by:
\begin{equation}
   \mathcal{L}_{u}=\frac{1}{N^u}\sum_{i=1}^{N^{u}}\frac{1}{H \times W}\sum_{j=1}^{H \times W}\mathcal{L}_{ce}(\hat{y}_{i,j}^{u}, y_{i,j}^{p})\cdot\mathds{1}[y_{i,j}^{p}\ne\varnothing],
\end{equation}
\noindent
where $\mathds{1}[\cdot]$ is the indicator function.

To address the problem of low contrast, LCC is introduced. More concretely, we first extract the feature maps of $x_{i}^{l}$, $x_{i}^{u}$ before last layer of the student network, denoted as $F_{i,j}^{l}, F_{i,j}^{u} \in \mathbb{R}^{D^{f}}$, where $D^{f}$ is the channel number of the feature maps. A projector $f_{p}(\cdot)$, which is a non-linear multiple-layer perception (MLP), and a bilinear interpolation operation are leveraged to map $F_{i,j}^{l}, F_{i,j}^{u}$ to the feature embeddings $z_{i,j}^{l},z_{i,j}^{u}$, formulated as:
\begin{align}
    z_{i,j}^{l}&=f_I(f_{p}(F_{i,j}^{l})),\\
    z_{i,j}^{u}&=f_I(f_{p}(F_{i,j}^{u})),
\end{align}
\noindent
where $f_I(\cdot)$ represents the bilinear interpolation operation. Here, $z_{i}^{l},z_{i}^{u}$ are interpolated to the same size as $x_{i}^{l},x_{i}^{u}$, which ensures that each pixel has a corresponding feature embedding.

Subsequently, the class embedding $\Bar{z}^{c}$ guided by the label data can be obtained by class-wise average pooling:
\begin{equation}
   \Bar{z}^{c}=\frac{\sum_{i,j}z_{i,j}^{l}\mathds{1}[y_{i,j}^{l}=c]}{\sum_{i,j}\mathds{1}[y_{i,j}^{l}=c]}.
\end{equation}
\noindent
It is noteworthy that we only leverage the feature embedding of label data $z_{i}^{l}$ to calculate the class embedding since there is potential noise in the pseudo-label $y_{i}^{p}$. Finally, both $z_{i,j}^{l}$, $z_{i,j}^{u}$ and $\Bar{z}^{c}$ are leveraged to calculate the LCC loss $\mathcal{L}_{lcc}$. With the loss function described above, our overall loss $\mathcal{L}$ can be defined as follows:
\begin{equation}
\mathcal{L}=\mathcal{L}_{sup}+\lambda_{1}\mathcal{L}_{u}+\lambda_{2}\mathcal{L}_{lcc},
\end{equation}
\noindent
where $\lambda_{1}$ and $\lambda_{2}$ are loss coefficients to adjust the weights of $\mathcal{L}_{u}$ and $\mathcal{L}_{lcc}$, respectively.

\subsection{LCC}
\label{sec LCC}
The goal of LCC is to mine low-contrast pixels and enhance the model's ability to distinguish these pixels via contrastive learning. Specifically, We first select low-contrast pixels from the perspectives of both class (\emph{i.e.}, tumor types) and tumor boundary. Each selected low-contrast pixel $x_{i,j}$'s corresponding feature embedding $z_{i,j}$ will be added to the low-contrast feature embedding set $\mathcal{H}$ (detail refer to section \ref{sec ICE} and \ref{sec BCE}). Then, given a feature embedding $z_{i,j} \in \mathcal{H}$ belongs to class $c$, we guide the model to learn discriminative features by pulling $z_{i,j}$ close to the class embedding $\Bar{z}^{c}$, while pushing it further away from the other class embeddings not belong to class $c$. Formally, let $\{z^{c}\}=\{z_{i,j}^{l}|y_{i,j}^{l}=c, z_{i,j}^{l} \in \mathcal{H}\} \cup \{z_{i,j}^{u}|y_{i,j}^{p}=c, z_{i,j}^{u} \in \mathcal{H}\}$ denote the set of low-contrast feature embeddings belong to class $c$. The loss of LCC can be defined as:
\begin{equation}
\begin{split}
\mathcal{L}_{lcc}=&-\frac{1}{K}\sum_{c}\frac{1}{|\{z^{c}\}|}\sum_{z_{i} \in \{z^{c}\}}\\
&\log\frac{\exp(z_i \cdot \Bar{z}^{c} / \tau)}{\exp(z_i \cdot \Bar{z}^{c} / \tau) + \sum_{z_j \in \mathcal{H} \backslash \{z^{c}\}}\exp(z_j \cdot \Bar{z}^{c} / \tau)},
\end{split}
\end{equation}
\noindent
where $\tau$ is a temperature parameter.
\subsubsection{ICE}
\label{sec ICE}
We first identify low-contrast pixels from the inter-class perspective. We measure the contrast of a pixel $x_{i,j}$ by calculating the class similarity $s^{c}_{i,j}$, which can be obtained by the similarity between $x_{i,j}$'s corresponding feature embedding $z_{i,j}$ and the class embeddings $\Bar{z}^{c}$ in the projection space. A lower class similarity suggests that the feature embedding of the pixel is likely located at the edge of the class distribution, indicating that the model has difficulty learning high-contrast features for the corresponding pixel. Formally, given a feature embedding $z_{i,j}$ belong to class $c$, its class similarity $s^{c}_{i,j}$ can be calculated by:
\begin{equation}
    s^{c}_{i,j}=z_{i,j} \cdot \Bar{z}^{c}.
\end{equation}

Here, we select the top $k\%$ feature embeddings with the lowest class similarity $s^{c}$ to add to the low-contrast pixel set $\mathcal{H}$.
\subsubsection{BCE}
\label{sec BCE}
We additionally identify low-contrast pixels from the boundary perspective. If, among the $h$ pixels closest to the position of $x_{i,j}$, there exists at least one pixel with a class label different from $y_{i,j}$, then $x_{i,j}$ is defined as a boundary pixel. For each boundary pixel $x_{i,j}$, we measure its boundary contrast by calculating the boundary feature similarity. Here, we define boundary feature similarity $s^{b}_{i,j}$ as the minimum similarity between $x_{i,j}$'s corresponding feature embedding $z_{i,j}$ and $x_{i,j}$ $h$ neighboring pixels' feature embeddings, formulated as:
\begin{equation}
    s^{b}_{i,j}=\mathop{\min}\limits_{j^{\prime} \in \mathcal{A}^{h}_{i,j}} z_{i,j} \cdot z_{i,j^{\prime}},
\end{equation}
\noindent
where $\mathcal{A}^{h}_{i,j}$ denotes the index of the $h$ pixels closest to $x_{i,j}$. A higher boundary similarity implies that the feature similarity between the boundary pixel and its surrounding pixels is greater, indicating a lower contrast for that pixel (\emph{i.e.}, the model struggles to classify that boundary pixel).

Following ICE, we select the top $k\%$ boundary pixels with the highest boundary similarity $s^{b}$ to add to the low-contrast pixel set $\mathcal{H}$.

\subsection{CDF}
\label{sec CDF}
In semi-supervised segmentation, it is essential to use a threshold to filter out low-confidence pseudo-labels and prevent confirmation bias. A common practice is to employ a fixed threshold. However, this approach has two significant drawbacks: 1) It can result in the loss of many valuable pseudo-labels during the early stages of training. 2) In endoscopic image segmentation, the issue of pixel imbalance is severe. Certain minority classes, such as benign tumor, account for only a small portion of the total pixels in all images. This often results in lower confidence for these minority classes, ultimately leading to extremely low utilization of pseudo-labels for minority classes. (see section \ref{ablation section}).

\begin{table*}[!tb]
  \begin{center}
    \caption{Segmentation results on FAHSYU-LC. \textbf{Bold} denotes the best performance, while \underline{underline} denotes the second.}

    \label{results on FAHSYU-LC}
    \newcolumntype{C}{>{\centering\arraybackslash}X}
    \begin{tabularx}{\textwidth}{l|CCCCC|CCCCC|CCCCC}
    \hline
      \multirow{2}{*}{Method} & \multicolumn{5}{c|}{10\%} & \multicolumn{5}{c|}{30\%} & \multicolumn{5}{c}{50\%}\\
      \cline{2-16}
      & IoU(M) & IoU(B) & mIoU & DSC & NSD & IoU(M) & IoU(B) & mIoU & DSC & NSD & IoU(M) & IoU(B) & mIoU & DSC & NSD \\
      \hline
      Supervised & 65.69 & 36.65 & 51.17 & 58.17 & 60.38 & 70.98 & 45.96 & 58.47 & 67.29 & 69.78 & 73.47 & 55.63 & 64.55 & 71.02 & 73.60 \\
      FixMatch & 70.20 & 46.44 & 58.32 & 66.79 & 69.21 & 73.93 & 49.00 & 61.46 & 70.18 & 72.71 & 73.62 & 54.38 & 64.00 & 70.08 & 72.67\\
      U$^{2}$PL & 68.33 & 46.68 & 57.50 & 65.13 & 67.54 & 72.31 & 50.81 & 61.56 & 71.02 & 73.66 & 73.93 & \underline{57.98} & 65.95 & 72.56 & 75.26 \\
      UniMatch & 70.08 & 47.88 & 58.98 & 66.77 & 69.28 & 73.67 & 49.11 & 61.39 & 69.84 & 72.36 & 74.42 & 55.13 & 64.77 & 71.06 & 73.61\\
      AugSeg & \underline{71.21} & \underline{48.83} & \underline{60.02} & \underline{68.06} & \underline{70.56} & \underline{74.22} & \underline{51.22} & \underline{62.72} & \underline{71.80} & \underline{74.42} & \underline{74.53} & 57.76 & \underline{66.14} & \underline{73.20} & \underline{75.89} \\
      CorrMatch & 67.09 & 41.92 & 54.50 & 64.77 & 67.08 & 71.11 & 44.37 & 57.74 & 68.34 & 70.80 & 71.21 & 48.14 & 59.67 & 67.63 & 70.07 \\
      Ours & \textbf{72.19} & \textbf{52.96} & \textbf{62.57} & \textbf{70.49} & \textbf{73.13} & \textbf{75.75} & \textbf{56.51} & \textbf{66.13} & \textbf{75.54} & \textbf{78.38} & \textbf{76.23} & \textbf{61.56} & \textbf{68.89} & \textbf{75.82} & \textbf{78.58} \\
      \hline
    \end{tabularx}
  \end{center}
\end{table*}

Inspired by \cite{wang2023freematch}, we design a confidence-based dynamic filter (\emph{i.e.}, CDF) for semi-supervised endoscopic image segmentation. To improve the utilization of pseudo-labels during the early stages of training, we first introduce a dynamic global threshold $\mathcal{T}^{g}$. This dynamic global threshold is determined by the average confidence of the predictions, represented as $\alpha^{g}$, which can be defined as follows:
\begin{equation}
\alpha^{g}=\frac{1}{K}\sum_{c}\frac{\sum_{i,j}\max(\hat{y}_{i,j}^{u})^{u}\mathds{1}[y_{i,j}^{p}=c]}{\sum_{i,j}\mathds{1}[y_{i,j}^{p}=c]}.
\end{equation}
To ensure that the dynamic global threshold remains stable and does not fluctuate significantly with the prediction confidence across different batches, we use EMA to update the dynamic global threshold $\mathcal{T}_{t}^{g}$, formulated as:

\begin{equation}
\mathcal{T}_{t}^{g}=\left\{
	\begin{aligned}
	  & \mathcal{T}_{init}^{g}, &\quad t=0,\\
	& \lambda \cdot \mathcal{T}_{t-1}^{g} + (1 - \lambda) \cdot \alpha^{g}, &\quad t>0,\\
	\end{aligned}
	\right
	.
\end{equation}
\noindent
 where $t$ represents the $t$-th training time step and $\lambda$ is a EMA parameter. Here, $\mathcal{T}_{init}^{g}$ denotes the initialization of the dynamic global threshold, which is set to 0.85.

Subsequently, to alleviate the issue of low utilization of pseudo-labels for minority classes, we introduce a local threshold $\mathcal{T}^{l}(c)$ to adjust the thresholds for class $c$. The local threshold $\mathcal{T}^{l}(c)$ is determined by the average confidence of each class. Similar to the global threshold, we leverage EMA to update the local threshold, defined as:

\begin{equation}
\mathcal{T}_{t}^{l}(c)=\left\{
	\begin{aligned}
	  & \frac{1}{K}, & \quad t=0,\\
	& \lambda \cdot \mathcal{T}^{l}_{t-1}(c) + (1 - \lambda) \cdot \alpha^{l}(c), & \quad t>0,\\
	\end{aligned}
	\right
	.
\end{equation}
\noindent
where $\alpha^{l}(c)$ is the average confidence of class $c$, which can be defined as:
\begin{equation}
\alpha^{l}(c)=\frac{\sum_{i,j}\max(\hat{y}_{i,j}^{u})^{u}\mathds{1}[y_{i,j}^{p}=c]}{\sum_{i,j}\mathds{1}[y_{i,j}^{p}=c]}.
\end{equation}

Finally, based on the global threshold, we adjust the dynamic threshold for each class using the corresponding local threshold. Integrating the global and local thresholds, the dynamic threshold $\mathcal{T}(c)$ for class $c$ can be defined as follows:
\begin{equation}
    \mathcal{T}(c)=\mathcal{T}^{g}(\frac{\mathcal{T}(c)^{l}}{\mathop{\max}\limits_{c} (\mathcal{T}(c)^{l})})^{\gamma},
\end{equation}
\noindent
where $\gamma$ is a hyperparameter, controlling the impact of the local threshold.
 
\section{Experiments}
\label{section:experiments}

\subsection{Experiment setup}
\subsubsection{Dataset}
The experiments are conducted on three endoscopic image datasets: one proprietary dataset of FAHSYU-LC, and two public datasets of Kvasir-SEG and CVC-ClinicDB.
\begin{itemize}
\item[$\bullet$] FAHSYU-LC. It contains 12,323 RGB laryngoscopic images collected from 2019 to 2021 (IRB No.[2023]755), including 5,144 images captured in narrow-band imaging (NBI) mode and 7,179 images captured in white-light imaging (WLI) mode. There are three categories: normal tissue, benign tumor, and malignant tumor. Among them, 3,642 images contain the benign tumor category, and 8,681 images contain the malignant tumor category. The training and validation sets have been split chronologically, with the training set consisting of 11,322 images and the validation set containing 1,000 images.

\item[$\bullet$]  Kvasir-SEG and CVC-ClinicDB. They are 2 datasets for polyp segmentation. Following previous work \cite{fan2020pranet,yue2022boundary,wu2023acl}, a total of 1450 images, including 900 from Kvasir-SEG and 550 from CVC-ClinicDB are used for training. The rest of the 162 images, including 100 from Kvasir-SEG and 62 from CVC-ClinicDB are utilized for evaluation. 
\end{itemize}
\subsubsection{Evaluation}
To evaluate the performance of our method in the semi-supervised setting, we conduct experiments under $10\%$, $30\%$, and $50\%$ partition protocols (\emph{i.e.}, only $10\%$, $30\%$, and $50\%$ of labeled data are available). We adopt the mean of intersection of union (mIoU), dice similarity coefficient (DSC), and normalized surface distance (NSD) as our evaluation metric. Additionally, on FAHSYU-LC, to evaluate the model's segmentation performance on different tumor types, we also report the class intersection of union (IoU) for each category. Here, "IoU(M)" represents the IoU for malignant tumor and "IoU(B)" represents the IoU for benign tumor. On Kvasir-SEG and CVC-ClinicDB, we only report the mIoU, DSC, and NSD since there are only two categories (\emph{i.e.}, background and tumor).

\subsubsection{Implementation details}
Experiments are conducted on a computing server equipped with an NVIDIA GeForce RTX 4090 GPU offering 24GB in total of video memory and an Intel(R) Core(TM) i9-10900X CPU providing 128GB of system memory. We leverage DeepLabV3+ with ResNet-50 pre-trained on ImageNet as the segmentation model. SGD optimizer with weight decay is applied in training and the initial learning rate is set to $1 \times 10^{-3}$. The loss coefficients $\lambda_{1}$ and $\lambda_{2}$ are empirically set as 0.5 and 0.1 respectively. The hyperparameters $k$, $h$, and $\gamma$ are set as 30, 64, and 0.25 respectively. The temperature parameter $\tau$ is set as 0.1. For FAHSYU-LC, the image is cropped to $321 \times 321$, and the model is trained for 100 epochs with a batch size of 8. For Kvasir-SEG and CVC-ClinicDB, the image is cropped to $384 \times 384$ and the model is trained for 80 epochs with a batch size of 8.
\subsection{Comparison with state-of-the-art methods}
We compare the proposed LoCo with five state-of-the-art algorithms: FixMatch \cite{sohn2020fixmatch}, U$^{2}$PL \cite{wang2022semi}, UniMatch \cite{yang2023revisiting}, AugSeg \cite{zhao2023augmentation}, and CorrMatch \cite{sun2024corrmatch}. For a fair comparison, all methods adopt DeepLabv3+ \cite{chen2018encoder} with ResNet50 \cite{ResNet} as the segmentation model.

Table \ref{results on FAHSYU-LC} reports the experiment results on FAHSYU-LC. The results show that our method consistently outperforms other state-of-the-art methods. Under $10\%$, $30\%$, and $50\%$ label partitions, our method outperforms the previous best by $2.55\%$, $3.41\%$, and $2.75\%$ in terms of mIoU. Moreover, our method brings greater improvements on minority classes such as benign, which surpasses the previous best approach by $4.13\%$, $5.29\%$, and $3.58\%$ on each split.

Table \ref{results on Kvasir-SEG} shows the comparison results with the state-of-the-art methods on Kvasir-SEG. It is clear that our method consistently outperforms the second-best method by $2.57\%$, $1.61\%$, and $1.83\%$ on Kvasir-SEG. Table \ref{results on CVC-ClinicDB} shows the comparison results on CVC-ClinicDB, our method archives the mIoU of $83.81\%$, $88.49\%$, and $92.05\%$ under $10\%$, $30\%$, and $50\%$ partition protocols, which is on par with the state-of-the-art methods.

\begin{table}[!tb]
  \begin{center}
    \caption{Segmentation results on Kvasir-SEG. \textbf{Bold} denotes the best performance, while \underline{underline} denotes the second.}

    \label{results on Kvasir-SEG}
    \newcolumntype{C}{>{\centering\arraybackslash}X}
    \newcolumntype{L}[1]{>{\raggedright\arraybackslash}p{#1}}
    \begin{tabularx}{0.5\textwidth}{L{0.05\textwidth}CCCCCCCCC}
    \hline
      \multirow{2}{*}{Method} & \multicolumn{3}{c}{10\%} & \multicolumn{3}{c}{30\%} & \multicolumn{3}{c}{50\%}\\
      \cline{2-10} & mIoU & DSC &  \multicolumn{1}{C}{NSD} & mIoU & DSC & \multicolumn{1}{C}{NSD} & mIoU & DSC & \multicolumn{1}{C}{NSD} \\
      \hline
        Supervised & 74.74 & 76.69 & 78.83 & 83.38 & 82.66 & 84.73 & 83.79 & 83.82 & 85.93 \\
        FixMatch & 83.05 & 82.38 & 84.44 & 84.57 & 83.99 & 86.02 & 84.60 & 84.05 & 86.14 \\
        U$^{2}$PL & 82.38 & 81.89 & 83.83 & 84.36 & 84.35 & 86.44 & \underline{85.40} & \underline{85.35} & \underline{87.40} \\
        UniMatch & 83.16 & 83.33 & 85.41 & 82.37 & 83.86 & 86.00 & 84.41 & 84.53 & 86.55 \\
        AugSeg & \underline{83.61} & \underline{84.39} & \underline{86.42} & \underline{85.16} & \underline{84.44} & \underline{86.53} & 85.27 & 84.50 & 86.44 \\
        CorrMatch & 81.18 & 81.77 & 83.83 & 84.38 & 83.97 & 86.18 & 85.14 & 83.67 & 85.76 \\
        Ours & \textbf{86.18} & \textbf{84.92} & \textbf{86.85} & \textbf{86.77} & \textbf{86.03} & \textbf{87.99} & \textbf{87.23} & \textbf{86.18} & \textbf{88.08} \\
      \hline
    \end{tabularx}
  \end{center}
\end{table}

\begin{table}[!tb]
  \begin{center}
    \caption{Segmentation results on CVC-ClinicDB. \textbf{Bold} denotes the best performance, while \underline{underline} denotes the second.}

    \label{results on  CVC-ClinicDB}
    \newcolumntype{C}{>{\centering\arraybackslash}X}
    \newcolumntype{L}[1]{>{\raggedright\arraybackslash}p{#1}}
    \begin{tabularx}{0.5\textwidth}{L{0.05\textwidth}CCCCCCCCC}
    \hline
      \multirow{2}{*}{Method} & \multicolumn{3}{c}{10\%} & \multicolumn{3}{c}{30\%} & \multicolumn{3}{c}{50\%}\\
      \cline{2-10} & mIoU & DSC &  \multicolumn{1}{C}{NSD} & mIoU & DSC & \multicolumn{1}{C}{NSD} & mIoU & DSC & \multicolumn{1}{C}{NSD} \\
      \hline
        Supervised & 76.23 & 61.24 & 65.20 & 82.01 & 72.49 & 76.43 & 88.51 & \underline{81.15} & \underline{85.01} \\
        FixMatch & 82.63 & 75.35 & 79.06 & 86.08 & 78.32 & 81.85 & 84.68 & 76.80 & 80.42 \\
        U$^{2}$PL & 83.51 & \underline{76.58} & \underline{80.03} & 86.92 & 80.88 & 84.43 & 87.64 & 80.77 & 84.58 \\
        UniMatch & 82.66 & 74.11 & 77.66 & 86.00 & 78.34 & 81.89 & 84.87 & 75.43 & 79.47 \\
        AugSeg & \textbf{84.13} & \textbf{78.49} & \textbf{81.81} & \underline{86.95} & \underline{81.00} & \underline{84.89} & \underline{89.57} & 80.12 & 83.45 \\
        CorrMatch & 80.47 & 71.78 & 75.88 & 83.60 & 76.28 & 80.35 & 83.63 & 75.10 & 79.16 \\
        Ours & \underline{83.81} & 76.31 & 79.67 & \textbf{88.48} & \textbf{83.73} & \textbf{87.29} & \textbf{92.05} & \textbf{85.16} & \textbf{88.72} \\
      \hline
    \end{tabularx}
  \end{center}
\end{table}

\subsection{Ablation Studies}
\label{ablation section}

\begin{table*}[tb]
  \begin{center}
    \caption{Ablation results of LoCo on FAHSYU-LC under 10\% label partition.}
    \label{ablation table}
    \begin{tabular}{c|ccccc|ccccc}
    \hline
    
       \makebox[0.05\textwidth][c]{Variant} & \makebox[0.04\textwidth][c]{$\mathcal{L}_{sup}$} & \makebox[0.04\textwidth][c]{$\mathcal{L}_{u}$} & \makebox[0.04\textwidth][c]{CDF} & \makebox[0.04\textwidth][c]{BCE} & \makebox[0.04\textwidth][c]{ICE} & \makebox[0.06\textwidth][c]{IoU(M)} & \makebox[0.06\textwidth][c]{IoU(B)} & \makebox[0.06\textwidth][c]{mIOU} &
       \makebox[0.06\textwidth][c]{DSC} &
       \makebox[0.06\textwidth][c]{NSD} \\
    \hline
    M1 & \checkmark & & & & & 65.69 & 36.65 & 51.17 & 58.17 & 60.38 \\
    M2 & \checkmark & \checkmark & & & & 70.42 & 45.68 & 58.05 & 66.76 & 69.28\\
    M3 & \checkmark & \checkmark & \checkmark &  & & 70.18 & 49.04 & 59.61 & 67.46 & 69.99 \\
    M4 & \checkmark & \checkmark & & \checkmark & & 69.91 & 47.49 & 58.70 & 67.35 & 69.84 \\
    M5 & \checkmark & \checkmark & & & \checkmark & 71.85 & 51.43 & 61.64 & 
70.00 & 72.61 \\
    M6 & \checkmark & \checkmark & & \checkmark & \checkmark & 72.04 & 52.35 & 62.20 & 70.47 & 73.06\\ 
    M7 & \checkmark & \checkmark & \checkmark & \checkmark & \checkmark & \textbf{72.19} & \textbf{52.96} & \textbf{62.57} & \textbf{70.49} & \textbf{73.13} \\ 
      \hline
    \end{tabular}
  \end{center}
\end{table*}

Table \ref{ablation table} presents the ablation results of the LoCo method on the FAHSYU-LC dataset under a 10\% label partition. It compares the performance of seven distinct model variants, labeled M1 through M7, as detailed below:

\begin{itemize}
    \item \textbf{M1}: The supervised baseline, utilizing only labeled data for model training.
    \item \textbf{M2}: Builds on M1 by adding an additional loss function, $\mathcal{L}_{u}$, specifically for unlabeled data.
    \item \textbf{M3}: Enhances M2 with the integration of the proposed CDF module.
    \item \textbf{M4}: Modifies M2 by incorporating BCE.
    \item \textbf{M5}: Also modifies M2, but by incorporating ICE instead.
    \item \textbf{M6}: Combines the approaches of M4 and M5, incorporating both BCE and ICE into M2.
    \item \textbf{M7}: Represents the full LoCo model, which integrates the CDF into M6.
\end{itemize}
Table \ref{ablation table} highlights several key findings regarding the performance of the model variants. M1 serves as the supervised baseline, utilizing only labeled data, and achieves a malignant IoU of 65.69\%, a benign IoU of 36.65\%, and a mIoU of 51.17\%. M2 enhances these metrics by incorporating an additional loss function, $\mathcal{L}_{u}$, for unlabeled data. Building on M2, M3 introduces the proposed CDF module, leading to further improvements across all metrics. M4 and M5 enhance M2 by integrating BCE and ICE, respectively, resulting in noticeable gains. M6 combines both BCE and ICE with M2, yielding even better outcomes. Finally, M7 represents the complete LoCo model, which integrates the CDF module into M6, achieving the highest performance overall, with a malignant IoU of 72.19\%, a benign IoU of 52.96\%, a mIoU of 62.57\%, a DSC of 70.49\%, and an NSD of 73.06\% among all variants.

\subsection{Effectiveness of ICE}
\begin{figure}[tb]
\centerline{\includegraphics[width=\linewidth]{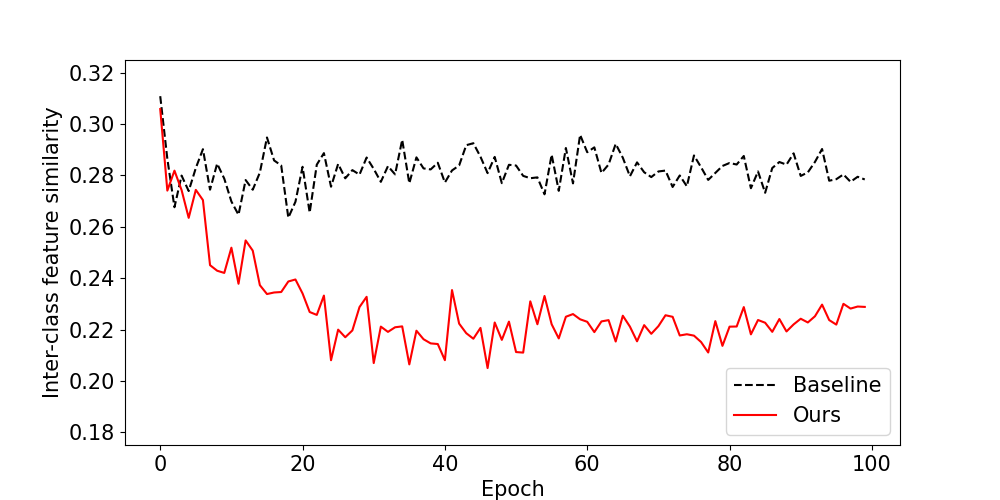}}
\caption{Visualization of the effectiveness of ICE. It is evident that LoCo, when combined with ICE, demonstrates superior inter-class feature similarity compared to the baseline method (\emph{i.e.}, M1).}
\label{fig ICE}
\end{figure}

Fig. \ref{fig ICE} presents a visualization of the effectiveness of ICE compared to a baseline (\emph{i.e.}, M1) over the course of training epochs. The plot shows the inter-class feature similarity, which measures the average similarity between the feature embeddings and the class embeddings of different categories. The solid line represents the results for the LoCo, which incorporates ICE. As the number of training epochs increases, the inter-class embedding similarity for LoCo decreases, indicating that the LoCo can progressively learn to extract higher-contrast features that are easier to distinguish between different classes. In contrast, the dashed line represents the results for the baseline without ICE. This line consistently remains higher than the LoCo line, suggesting that without ICE, the similarity between features of different classes remains relatively high compared to the LoCo method. This makes it more challenging for the model to effectively classify the corresponding features. 

\subsection{Effectiveness of BCE}
\begin{figure}[tb]
\centerline{\includegraphics[width=0.9\columnwidth]{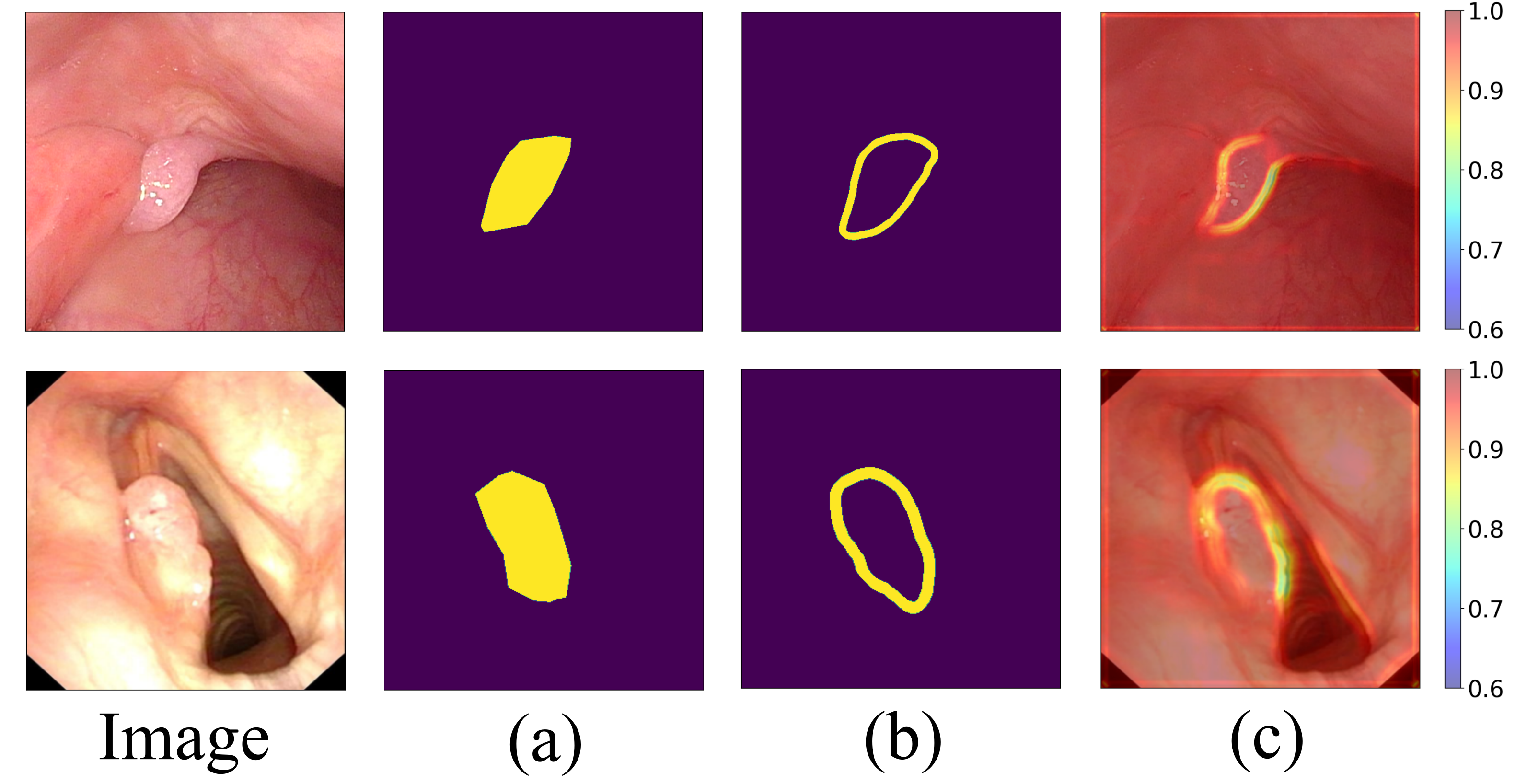}}
\caption{Visualization of boundary captured by BCE: (a) Ground truth, (b) Predicted boundary, and (c) Boundary feature similarity.}
\label{fig BCE1}
\end{figure}

\begin{figure}[tb]
\centerline{\includegraphics[width=1.\columnwidth]{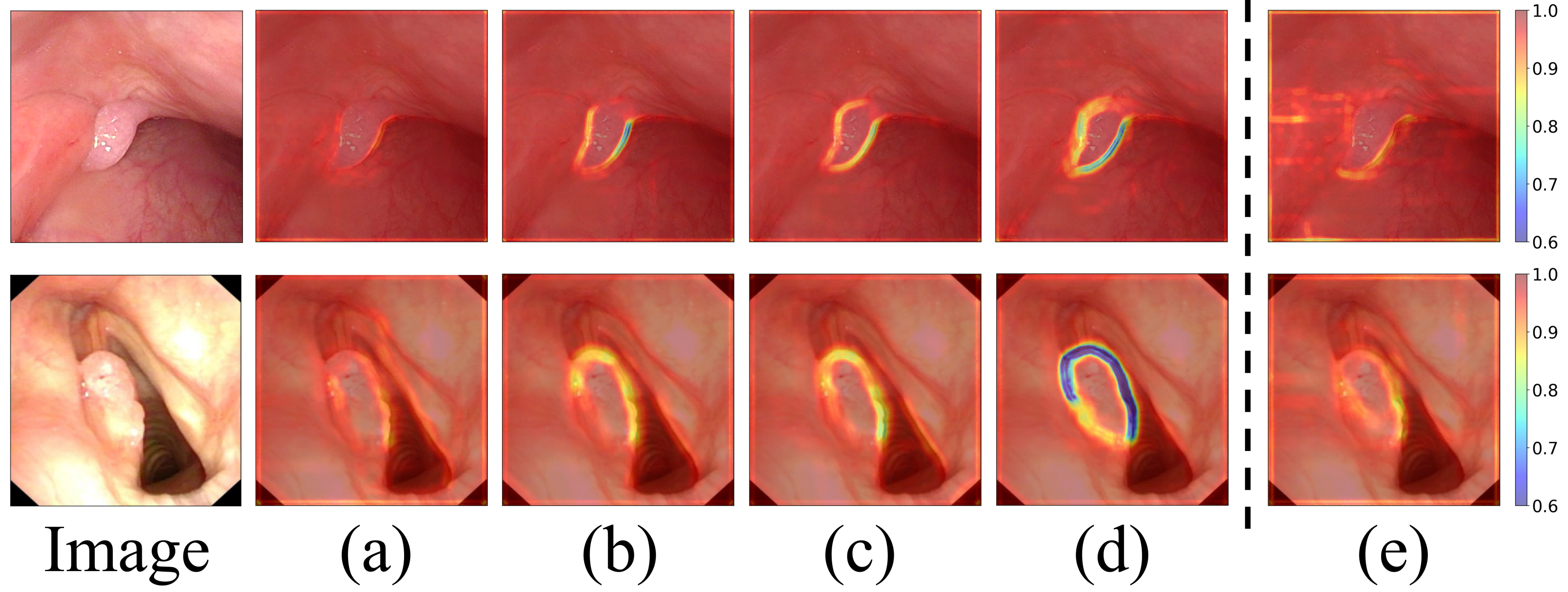}}
\caption{Visualization of boundary feature similarity at different epochs: (a) Epoch 1, (b) Epoch 5, (c) Epoch 10, (d) The best epoch, and (e) The best epoch of baseline (M1).}
\label{fig BCE2}
\end{figure}

Fig. \ref{fig BCE1} presents a visualization analysis of the boundaries captured by BCE. Plot (a) displays the ground truth, while plot (b) illustrates the predicted boundary. Plot (c) depicts the boundary feature similarity, which can effectively identify low-contrast boundary pixels. Among the boundary pixels, a higher boundary feature similarity indicates that the boundary pixel is more difficult for the model to distinguish. Fig. \ref{fig BCE2} (a-e) further explores the boundary feature similarity across different training epochs. As training progresses from (a) to (d), the boundary feature similarity surrounding the tumor contours decreases, indicating that the model is becoming more adept at distinguishing low-contrast boundary pixels. In contrast, plot (e) illustrates the scenario when BCE is not applied; here, the boundary feature similarity around the tumor edges remains elevated, suggesting that the model struggles to effectively learn boundary features without the enhancement provided by BCE. 

\subsection{Effectiveness of CDF}
\begin{figure}[!tb]
\centerline{\includegraphics[width=\linewidth]{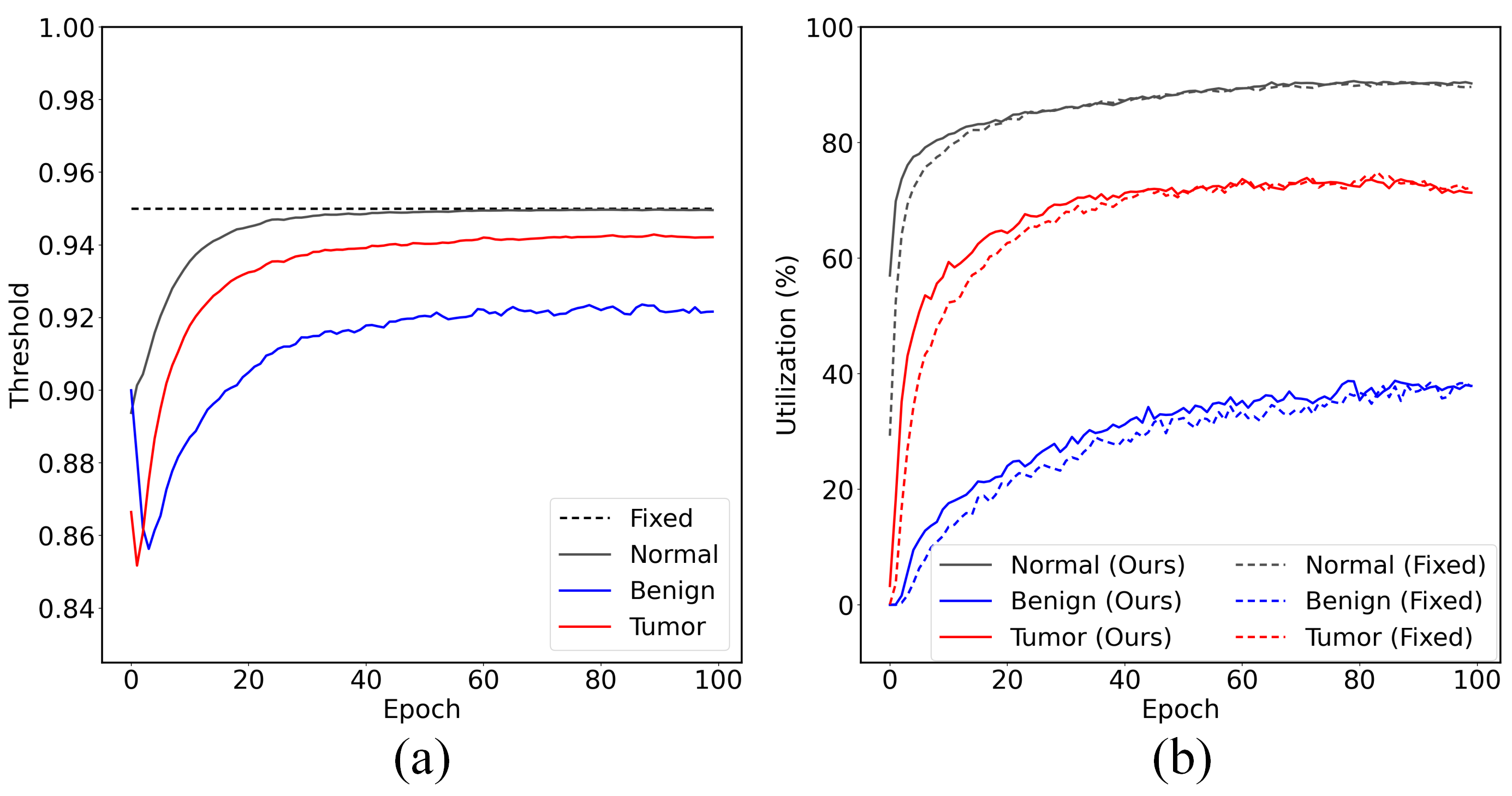}}
\caption{Visualization of the effectiveness of CDF: (a) The threshold for filtering pseudo-labels, and (b) The utilization of pseudo-labels. Here, "Fixed" means the threshold value is set to 0.95.}
\label{fig CDF}
\end{figure}
Fig. \ref{fig CDF} provides a visualization analysis of the effectiveness of the proposed CDF in filtering pseudo-labels for the different-classes segmentation task. 
The left plot (a) shows the threshold values for filtering pseudo-labels over different epochs. The lines represent the threshold values for different classes - normal, benign, and tumor. The "Fixed" line represents a fixed threshold value of 0.95, while the other lines show the dynamic threshold values determined by the CDF method. The right plot (b) shows the utilization rate of the pseudo-labels over different epochs for the same classes. The solid lines represent the utilization rates when using the CDF method, while the dashed lines represent the utilization rates when using a fixed threshold of 0.95. The key observation is that the CDF method (solid lines) significantly improves the utilization of pseudo-labels, especially for the underrepresented "benign" class, compared to using a fixed threshold (dashed lines). This indicates that the proposed CDF approach effectively addresses the challenge of low pseudo-label utilization for minority classes.

\subsection{Visualization Analysis under Various Label Partitions}
\begin{figure}[tb]
\centerline{\includegraphics[width=1.\columnwidth]{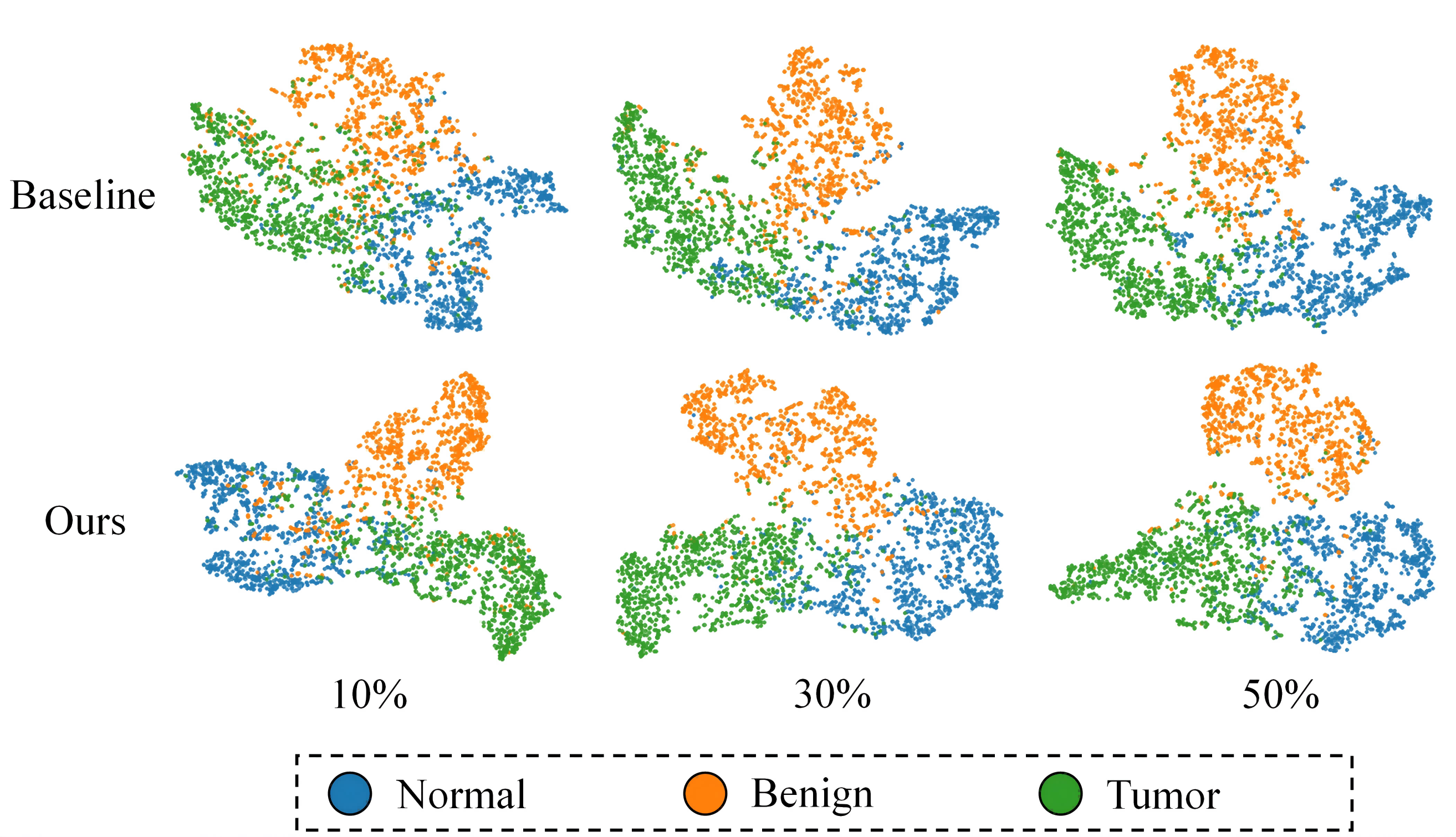}}
\caption{t-SNE Visualization of the baseline (M1) and LoCo under $10\%$, $30\%$, and $50\%$ label partitions, respectively.}
\label{fig TSNE}
\end{figure}

Fig. \ref{fig TSNE} provides a t-SNE visualization of the supervised baseline (M1) and the proposed LoCo under different label partitions of 10\%, 30\%, and 50\%. The visualizations show the distribution of normal, benign, and tumor samples in the feature space. Under the 10\% label partition, LoCo is able to extract more concentrated intra-class features compared to the baseline. This indicates that LoCo can learn more discriminative features from the limited labeled data. Under the 30\% and 50\% label partitions, both the baseline and LoCo are able to learn relatively accurate features. However, the features learned by LoCo exhibit clearer decision boundaries and better separability between the different classes, demonstrating that the proposed method enables the model to learn more discriminative features in the endoscopic images. Overall, the t-SNE visualizations validate the effectiveness of the proposed LoCo method, especially under the challenging 10\% label partition scenario, where LoCo outperforms the baseline in learning more discriminative features from the limited labeled data.

\begin{figure*}[tb]
\centerline{\includegraphics[width=0.85\linewidth]{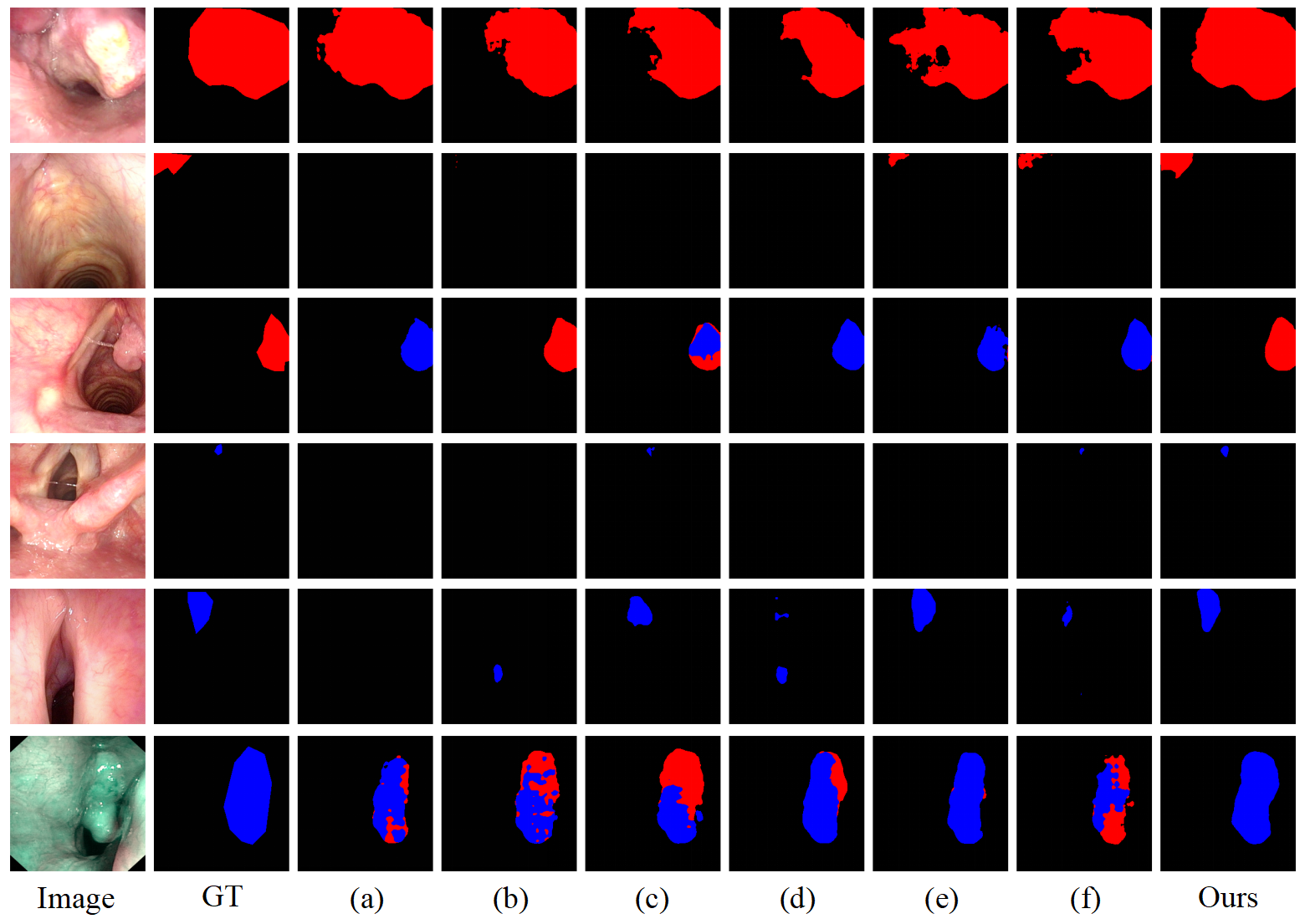}}
\caption{Visualization of segmentation results on FAHSYU-LC with $30\%$ labeled data. The segmentation results are presented for various methods as follows: (a) Supervised, (b) CorrMatch, (c) FixMatch, (d) UniMatch, (e) U$^{2}$PL, and (f) AugSeg. GT refers to the ground truth mask. The term "GT" denotes the ground truth mask, while "Ours" means the proposed LoCo. In the visual representation, the red mask indicates malignant tumor tissue, whereas the blue mask represents benign tumor tissue.}
\label{fig visualization}
\end{figure*}
\subsection{Segmentation Visualization Analysis}
In this section, Fig. \ref{fig visualization} illustrates the segmentation results on the proprietary FAHSYU-LC dataset with 30\% labeled data, using various semi-supervised learning methods. The methods shown are: (a) Supervised, (b) CorrMatch, (c) FixMatch, (d) UniMatch, (e) U$^{2}$PL, (f) AugSeg, and the proposed LoCo (the last column of Fig. \ref{fig visualization}). The segmentation results indicate that most methods can accurately segment large tumors, as shown in the first row of Fig. \ref{fig visualization}. However, these methods struggle to capture small tumors or differentiate low-contrast pixel tissues among malignant, benign, and tissue, as shown in rows 2 to 6. Among these recent advanced semi-supervised models, AugSeg appears to perform the best, but still exhibits a gap between the ground truth and predicted segmentation masks. In comparison, the proposed LoCo method demonstrates the best performance, with a high degree of overlap between the ground truth and predicted segmentation masks, even for small and low-contrast tumor tissues. This suggests that the LoCo is a promising semi-supervised approach for accurate tumor segmentation, particularly in challenging cases involving small or low-contrast lesions.

\section{Conclusion}
\label{section:conclusion}
In this study, we introduce a cutting-edge semi-supervised segmentation framework called LoCo, designed for endoscopic image segmentation through low-contrast-enhanced contrastive learning. Our approach features two innovative strategies aimed at improving the distinctiveness of low-contrast pixels: inter-class contrast enhancement (\emph{i.e.}, ICE) and boundary contrast enhancement (\emph{i.e.}, BCE). These strategies empower the LoCo to effectively distinguish between malignant tumors, benign tumors, and normal tissues. Furthermore, we implement a confidence-based dynamic filter to optimize the selection of pseudo-labels, enhancing the representation of minority classes. Comprehensive experiments conducted on two public datasets, along with a substantial proprietary dataset collected over three years, reveal that LoCo achieves state-of-the-art performance, significantly surpassing previous methods. The effectiveness of the proposed techniques is further verified through visualization analyses. The visualization results provide compelling evidence that ICE and BCE strategies are crucial components of the LoCo framework. These techniques empower the model to learn distinctive features that are essential for accurate segmentation of endoscopic images, even when faced with limited labeled data. Overall, the comprehensive evaluations validate the effectiveness of the LoCo framework and its potential to significantly advance the state-of-the-art in endoscopic image segmentation.

\section*{Acknowledgment}
We would like to extend our heartfelt thanks to Yixuan Yuan and Yongcheng Li for their invaluable suggestions on enhancing the quality of this manuscript. 



\newpage

\begin{IEEEbiography}[{\includegraphics[width=1in,height=1.25in,clip,keepaspectratio]{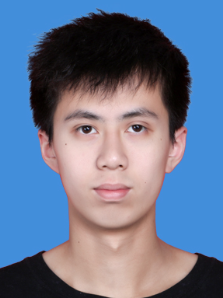}}]{Lingcong Cai} is currently studying toward the Bachelor's degree from Shenzhen Technology University of Computer Science and Technology, Shenzhen, China, since 2021. His research interests include deep learning and computer vision, especially focusing on semantic segmentation.
\end{IEEEbiography}

\vspace{11pt}

\begin{IEEEbiography}[{\includegraphics[width=1in,height=1.25in,clip,keepaspectratio]{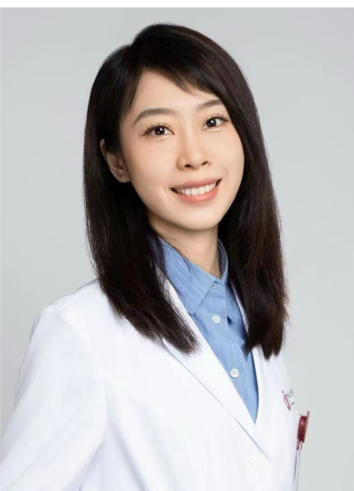}}]Yu Li (Member, IEEE) is a physician and assistant research fellow at the First Affiliated Hospital of Sun Yat-sen University. She earned her Ph.D. in Clinical Medicine (Otorhinolaryngology) from Sun Yat-sen University. Her research focuses on developing intelligent diagnostic models for head and neck tumors and exploring molecular biomarkers for tumor diagnosis. In recent years, She has published multiple internationally recognized articles on head and neck tumors in esteemed journals, including the Journal of Translational Medicine, Molecular Carcinogenesis, and the Journal of Otolaryngology-Head \& Neck Surgery. Additionally, she has applied for over ten patents related to her research.
\end{IEEEbiography}

\vspace{11pt}

\begin{IEEEbiography}[{\includegraphics[width=1in,height=1.25in,clip,keepaspectratio]{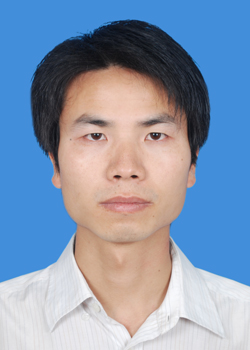}}]{Xiaomao Fan} (Member, IEEE) is an associate professor of Shenzhen Technonlogy University. He graduated from the University of Chinese Academy of Sciences with a Ph.D. degree in Applied Computer Technology. His research focuses on deep learning and medical big data mining. In recent years, he has published multiple papers in prestigious international journals such as IEEE Transactions on Industrial Informatics, IEEE Journal of Biomedical and Health Informatics, IEEE Transactions on Big Data, and IEEE Transactions on Network and Service Management. He has applied for over 40 patents, with more than 20 granted, and leads serval projects.
\end{IEEEbiography}

\vspace{11pt}

\begin{IEEEbiography}[{\includegraphics[width=1in,height=1.25in,clip,keepaspectratio]{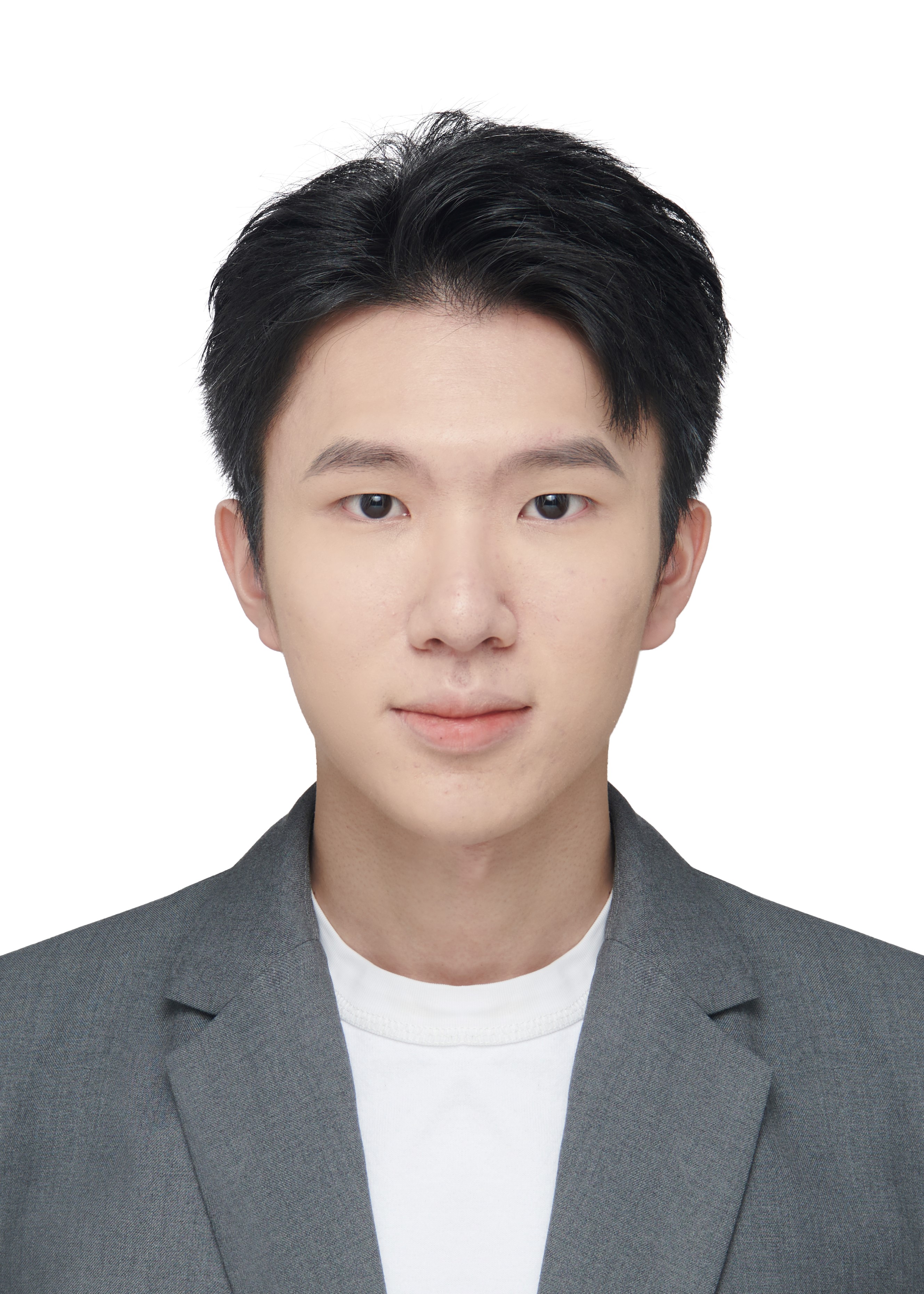}}]{Kaixuan Song} is currently studying toward the bachelor’s degree from Shenzhen Technology University of Computer Science and Technology,Shenzhen,China,since 2021.His research interests including deep learning,medical image classification and Semi-Semantic segmentation.
\end{IEEEbiography}

\vspace{11pt}

\begin{IEEEbiography}[{\includegraphics[width=1in,height=1.25in,clip,keepaspectratio]{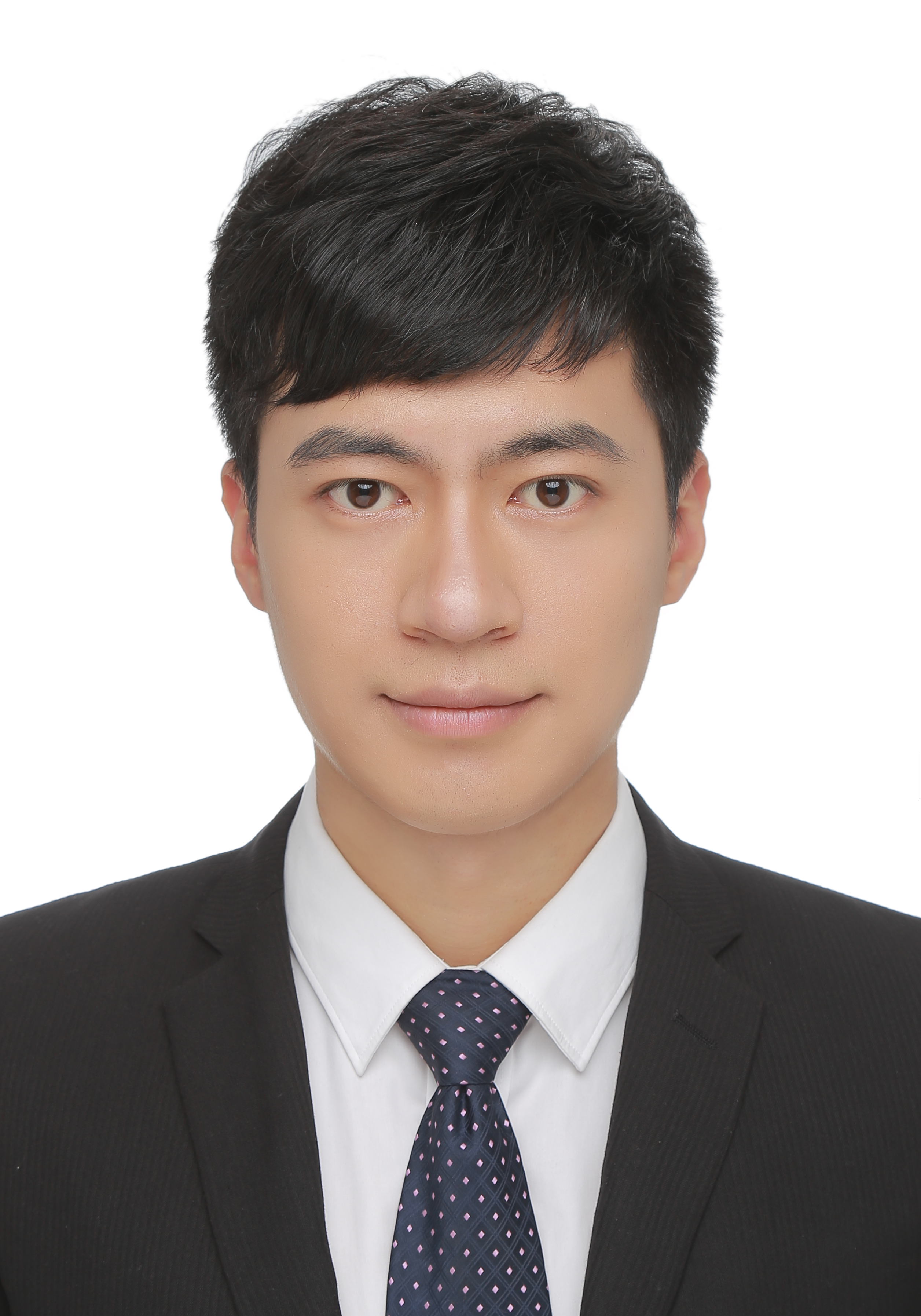}}]{Ruxin Wang} (Member, IEEE) received the PhD degree in operational research and cybernetics from the University of Chinese Academy of Sciences, Beijing, China, in 2017. He is currently an Associate Professor with the Shenzhen Institute of Advanced Technology, Chinese Academy of Sciences, Shenzhen, China. His current research interests include artificial intelligence in image processing, causal inference, machine learning, and pattern recognition. 
\end{IEEEbiography}

\vspace{11pt}

\begin{IEEEbiography}[{\includegraphics[width=1in,height=1.25in,clip,keepaspectratio]{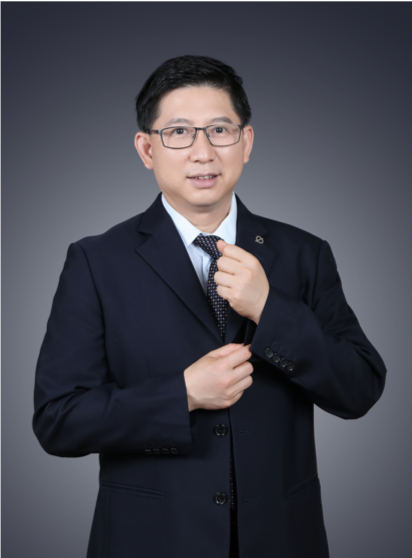}}]{Wenbin Lei} is the Director of the Department of Otorhinolaryngology at the First Affiliated Hospital of Sun Yat-sen University. He earned his Ph.D. in Otorhinolaryngology from Sun Yat-sen University. With over 20 years of clinical experience, he specializes in complex cases within otorhinolaryngology and head and neck malignancies. His research interests include tumor mechanisms, precision prevention, and the integration of digital health, virtual simulation, and big data AI in medicine. He has published numerous high-impact medical papers, holds over 10 patents, and has received prestigious awards, including the First Prize for Teaching Achievements and the "Master of Unique Skills" title from the Guangdong Medical Association. 

\end{IEEEbiography}

\vfill

\end{document}